# The end of Sleeping Beauty's nightmare

*Abstract*

The way a rational agent changes her belief in certain propositions/hypotheses in the light of new evidence lies at the heart of Bayesian inference. The basic natural assumption, as summarized in van Fraassen's Reflection Principle ([1984]), would be that in the *absence* of new evidence the belief should not change. Yet, there are examples that are claimed to violate this assumption. The apparent paradox presented by such examples, if not settled, would demonstrate the inconsistency and/or incompleteness of the Bayesian approach and without eliminating this inconsistency, the approach cannot be regarded as scientific.

The Sleeping Beauty Problem is just such an example. The existing attempts to solve the problem fall into three categories. The first two share the view that new evidence is absent, but differ about the conclusion of whether Sleeping Beauty should change her belief or not, and why. The third category is characterized by the view that, after all, new evidence (although hidden from the initial view) is involved.

My solution is radically different and does not fall in either of these categories. I deflate the paradox by arguing that the two different degrees of belief presented in the Sleeping Beauty Problem are in fact beliefs in two different propositions, i.e. there is no need to explain the (un)change of belief.

## 1 The Sleeping Beauty Problem

The Sleeping Beauty Problem is a paradox in probability theory that has recently received much attention in the literature. Sleeping Beauty (SB) undergoes the following experiment (the setup of which is known to her). She is put to sleep on Sunday evening. Then an experimentalist tosses a fair (50:50) coin. If the result is Heads, she is woken on Monday only. If the result is Tails, SB is woken twice, on Monday and on Tuesday. In addition, she is given a special drug that causes her to forget whether she was woken on the day before or not. Thus, when she is woken she does not know whether it is Monday or Tuesday. Each time she is woken she is asked to give her credence the coin landed Heads. The 'problem' is that the two following answers both seem to be valid[1]:

(a) On the one hand, on Sunday SB believed that the coin is fair and the probability that it will land Heads is ½. The fact that SB has been woken does not give her any new relevant information, because she knew all along that she is supposed to be woken at least once anyway. Thus, her credence ought to stay the same as on Sunday, i.e. 1/2.
(b) On the other hand, on a long series of trials the number of times she will be woken after the coin landed Tails is twice the number of times SB will be woken after the coin landed Heads. Thus, her answer should be 1/3.

---
[1] There are several versions of the problem and a few variations of how the experiment is conducted. These variations are, however, not crucial for the argument of the present article.

Two remarks should be made about these answers. First, the question posed to SB is a question about her credence upon awakening. This is not a question about the nature of the coin. Thus, any change of SB's credence (if any) must not be interpreted as a change of her opinion about the nature of the coin. Second, the plausibility of the answer (b) can be made more profound. There is no qualitative difference between the original formulation of the SB problem and its following modification. Suppose that SB is put to sleep on New Year's Eve. At 12pm the champagne is opened and the fair coin is tossed. Subsequently, if it lands Tails, SB is woken on every single day during the year, i.e. 365 times, if the coin lands Heads she is woken only once on one (arbitrary) day during the year. In this experiment, therefore, SB's choice will be between ½ and 1/366. One would find much harder to argue in favour of ½ in this case. With an appropriate modification of the experiment, the second number can take any rational value. It will be especially difficult to maintain the answer of ½ in the limit of that number going to zero.

## 2 The problem deflated

### 2.1 From contradiction to consistency

How can the problem be resolved? The existing approaches in the literature are characterized by a common assumption (explicit or implicit) that the reasoning (a) is in logical contradiction with the reasoning (b). Therefore, the usual efforts to solve the problem focus on demonstrating that, despite the apparent correctness of both (a) and (b), one of them is false (Weintraub [2004]). And there is a good deal of controversy amongst philosophers about which one! "Thirders" are trying to provide various explanations why SB ought to change her belief from ½ on Sunday to 1/3 on awakening. "Halfers", on the contrary, are trying to justify why her belief ought to stay the same, i.e. ½. Combining above with the view on the presence/absence of new evidence the existing strategies to solve the problem can be classified into three categories. The first two share the view that new evidence is absent, but differ about the conclusion of whether SB should change her belief or not, and why–"thirders" Elga ([2000]), Vaidman and Saunders ([2001]), Monton ([2002]), Hitchcock ([2004]) and "halfers" Lewis([2001]), Meacham ([2005]), White ([2006])[2]. The third category is characterized by the view that, after all, new evidence (although hidden from the initial view) is involved–"thirders" Dorr([2002])[3], Arntzenus ([2003])[4], Weintraub ([2004]), Horgan ([2004], [2007]).

---

[2] Strictly speaking, White does not explicitly states that he is a "halfer". He proposes a generalized version of the problem, which apparently poses a challenge for "thirders", in particular Elga-Dorr-Arntzenius arguments, but which does not pose any problems for "halfers". Though, Horgan ([2007]) denies that White's argument poses any problem for his approach.
[3] Dorr's argument was disputed by Bradley ([2003]).
[4] In his earlier article Arntzenius ([2002]) maintained a view that upon awakening SB should not have a definite belief at all due to her cognitive malfunction.

My approach is completely different. I question the very basic assumption that (a) contradicts (b) and deflate the problem by arguing that this contradiction is merely apparent. In other words, I show that there is no contradiction between SB's belief that the coin is fair, i.e. upon tossing on Sunday the probability of the coin to land Heads is equal ½, and her credence of 1/3 upon awakening that the result of coin tossing was Heads. Thus, on the one hand, I agree with "thirders" that her credence on awakening should be 1/3, but on the other hand, I show that there is no need for SB to *change* her belief. One belief does not contradict the other.

Moreover, this contradiction can be dissolved solely in the framework of standard probability theory. I show that no additional arguments going beyond basic probability theory such as appeals to the relevance of a rational agent's 'own temporal location' (Elga [2000]) or the difference between knowing all along that SB will be woken and knowing that she is woken *now* (Weintraub [2004]) are needed.

I will argue that both (a) and (b) are correct answers but to two *different* questions. SB is asked to give her credence, i.e. the degree of belief, in the particular value of physical or epistemic probability, of a certain *event*. However the phrase "the coin landed Heads" alone does not define that event completely. As I will discuss in detail in the course of this article, an experimental setup is necessary to describe an event. In the question posed to SB we implicitly assume that setup. This setup is wakening (and interviewing)[5]. Thus, the question posed to her is 'What is your credence the coin landed Heads under the setup of wakening?' It can be rephrased as 'What is your credence that this awakening is a Head-awakening under the setup of wakening?' And the correct answer should be (b). However, "the coin landed Heads" with a different setup will form a different event. In particular, "the coin landed Heads under the setup of coin tossing" would be a different event. If we asked SB 'What is your credence that the coin landed Heads under the setup of coin tossing?', then the correct answer would be (a) [SB still believes that the coin is fair]. At first sight, these two questions might seem similar, especially because there is a one-to-one logical cause-effect correspondence between them. Nevertheless they are not. Consequently, the corresponding probabilities are different as well. Thus, the source of the problem is that, although, the former question is the one that is posed to SB, one tends to confuse it with the latter question, thereby arriving at a paradox.

The aim of the rest of the article is to justify the argument presented in the preceding paragraph, and, in particular, to provide detailed explanation of why those two questions are different and why the two events associated with them are different as well.

The following notation will be used below:
P(A) – the probability of an event A,
P(A/B) – the conditional probability of an event *A* given that an event *B* has occurred.
Trial – a single run of the complete experiment starting with the act of coin tossing and ending with one or two awakenings.

---

[5] Note that including a setup in the definition of an event is different from the conditioning of credence of the event on evidence. SB does not receive any new evidence upon wakening, yet the credences are not the same, because the setups, and therefore the events, are different.

## 2.2 The inanimate version

In order to emphasize the objective mathematical character of this illusion and eliminate every possible psychological/subjective aspect often surrounding this paradox, I will present it with the following inanimate version of the problem. Let me replace the coin, the experimentalist and SB by the following automatic setting: an automatic device tosses a fair coin, if the coin lands 'Tails' the device puts two red balls in a box, if the coin lands 'Heads' it puts one green ball in the box. The device repeats this procedure a large number of times, N. As a result the box will be full of balls of both colours.

The device's task now is to determine the probability that if it picks up a ball from the box at random, this ball will be a green ball. The device may calculate this probability theoretically using the relative frequency definition of probability[6]. At large N the probability of picking up a green ball, P(green←box), is numerically very close to the ratio of the number of green balls to the total number of balls in the box:

$$P(green \leftarrow box) \approx N(green)/N(green+red). \qquad (1)$$

Because the coin is fair then in the long run it will land 'Heads' approximately N/2 times. Therefore, the number of green balls in the box will be approximately half of the total number of trials, N(green)≈N/2. Similarly, N(red)≈N. Thus, N(green+red)≈3N/2, and we obtain

$$P(green \leftarrow box) = 1/3. \qquad (2)$$

The device may, of course, verify this result experimentally simply by counting the numbers of balls in the box.

On the other hand, since the event 'Coin landed Heads', P(H), is necessarily followed by the event 'A (one) green ball is put in the box' (green→box) there is one-to-one correspondence between these two events, i.e.

$$P(green \rightarrow box) = P(green \rightarrow box\ /H)\ P(H)$$
$$= P(H/green \rightarrow box) P(green \rightarrow box) = P(H), \qquad (3)$$

according to Bayes' rule. Now we have arrived at a critical point. The core of the whole confusion is that we tend to regard 'A (one) green ball is *put in* the box' and 'A green ball is *picked out* from the box' as equivalent. Subsequently we combine (2) and (3), thereby (mistakenly) concluding that the probability of 'Coin landed Heads' is 1/3. But, of course, it is not!

The reason is that the event 'A green ball is put in the box' and the event 'A green ball is picked out from the box' are two *different* events, and therefore their probabilities are not necessarily equal. These two events are different because they are the subject to *different experimental setups*: one is the coin tossing, other is picking up a ball at random from the full box[7]. The probability to put a green ball in the box on each coin tossing trial is ½, but the probability to pick out a green ball from the full box at the end is 1/3! This might seem paradoxical, but there is no contradiction.

---

[6] Note that here as well as in the original statement of the paradox, as formulated by Elga ([2000]), the frequentist definition of probability is used in (b). In subsequent discussions, though, Elga ([2000]) and other authors based their arguments mainly on the application of the principle of indifference and on Bas van Fraassen's reflection principle, rather than on frequentist definition of probability. In this article I use the frequentist definition simply because it does the job perfectly. Moreover, the way I dissolve the problem implies that application of Bayesian methods will not lead to any contradictions as well.

So, how from the first ½ do we get that only 1/3 of all balls in the box are green, and therefore 1/3 as the probability to pick out a green one? Although the probability of a green ball being put in the box in each trial is ½, the *average number* of green balls which the device puts in the box on each trial is 1/3. That is why at the end the number of green balls in the box is half the number of red ones. Let me show this calculation in detail.

The total average number of balls which the device puts in the box on one trial is:

$$N_{av}[\text{green+red}\rightarrow\text{box}] = P(H)\cdot 1 + P(T)\cdot 2 = 3/2.$$

Here I take into account the fact that if the coin lands Tails then two red balls go into the box. Therefore,

$$N_{av}[\text{green}\rightarrow\text{box}] = P(H)\cdot 1 / N_{av}[\text{green+red}\rightarrow\text{box}] = 1/3.$$

### 2.3 Back to SB

Hopefully, now everyone is convinced that there is nothing wrong in believing in both propositions: the probability of a green ball being put in the box equals 1/2 and the probability of a green ball being picked out from the box equals 1/3. The case of SB should be no different.

Indeed, in a direct analogy with the experiment described above, SB's 'Head/Tail'-awakenings are just like those green/red balls. In the long run of N coin tossing trials there will be twice as many Tail-awakenings 'in the box' than Head-awakenings, where an awakening is the analogue of 'being picked out from the box'. The probability that 'This awakening is a Head-awakening *under the setup of wakening*' is 1/3. However, although a Head-awakening is necessarily preceded by the coin landing Heads, the probability of 'The coin landed Heads *under the setup of coin tossing*' is ½, i.e. on each coin toss the probability that one Head-awakening will be 'added to the box' is ½. Yet, the average number of Head-awakenings 'added to the box' on each coin tossing trial is 1/3. Thus everything is consistent. And the answer depends on what precisely do we mean by the question. If we mean 'This awakening is a Head-awakening *under the setup of wakening*', then SB's answer to our question should be 1/3, but if we mean 'The coin landed Heads *under the setup of coin tossing*', her answer should be 1/2.

After all these explanations, you might still wonder why the original question is so ambiguous. The two setups are very different from each other. So, why do we tend to confuse them if they are not mentioned in the question explicitly? One tends to assign the probability of 1/3 to the event 'The coin landed Heads' if one assumes 'The coin landed Heads *under the setup of wakening*' instead of '*under the setup of coin tossing*'. But what sense does it make to define such an event? As the Head-awakening is necessarily preceded by the coin landing Heads, the above formulation is equivalent to 'This awakening is a Head-awakening *under the setup of wakening*'. The interpretation of the latter is clear and was given in the preceding paragraph. The interpretation of the former version might be the following. To say that the awakening is a Head-awakening is to say that the last time the coin was tossed it had landed Heads, and the experimenter recorded this result (say by writing it on a piece of paper). To ask SB to give the probability of 'The coin landed Heads *under the setup of wakening*' means to ask her to give the probability of finding 'Heads' written now on that piece of paper.

---

[7] This emphasises the fact that an experimental setup or condition of an event is essential for its (event's) definition (Shaposhnikov [1987]). Even two similarly-looking events are different if they are subject to different conditions. Thus, we should define the two above events more precisely as 'A green ball is picked out from the box *under the setup that the device picks out a ball from the full box*' and 'A green ball is put in the box *under the setup that the device tosses a fair coin*'.

The answer 1/3 is often justified by an appeal to the betting approach to probability–e.g. see Hitchcock ([2004]). (Although, some authors, e.g. Bradley and Leitgeb ([2006]), dispute its applicability to SB problem). There is, obviously, no contradiction between the argument of this article and the betting approach, which corresponds to the setup of wakening. Indeed, the number of times SB gives a right answer relative to the number of times *she is asked* is what counts here. The number of times SB is asked equals the number of times she is woken. This is clearly the setup of wakening. (The drug makes each of these questionings independent of the others.)

### 3 Summary

The concept of an *event* is central and crucial in Probability Theory. The Sleeping Beauty Problem arises due to improper use of the notion of an *event*. The *setup* under which the event takes place must always be taken into account. If we do so, then we realize that the original question posed to SB can be interpreted in two different ways. The first interpretation is 'What is your credence that the coin landed Heads *under the setup of coin tossing*?', and the answer should be ½. The second interpretation is 'What is your credence that this awakening is a Head-awakening *under the setup of wakening*?', and the answer should be 1/3. Thus there is no paradox!

### Funding
The UK Engineering and Physical Sciences Research Council (EP/C528042/1).

### Acknowledgements
I would like to thank James Ladyman for very helpful and encouraging comments and support, Jeremy Butterfield for his valuable remark and useful corrections, and anonymous referee for her/his positive feedback and corrections. I am grateful to Lev Vaidman, who first pointed my attention to the Sleeping Beauty Problem, for constructive discussions.

*Berry Groisman*
*Centre for Quantum Computation*
*Department of Applied Mathematics and Theoretical Physics,*
*University of Cambridge,*
*Cambridge CB3 0WA, UK*
*b.groisman@damtp.cam.ac.uk*